\documentclass[12pt,journal,onecolumn]{IEEEtran}
\IEEEoverridecommandlockouts
\usepackage{cite}
\usepackage{amsmath}
\usepackage{amssymb,amsfonts}
\usepackage{algorithmic}
\usepackage{graphicx}
\usepackage{textcomp}
\usepackage{xcolor}
\usepackage{lipsum}
\usepackage{setspace}
\usepackage{flushend} 
\def\BibTeX{{\rm B\kern-.05em{\sc i\kern-.025em b}\kern-.08em
    T\kern-.1667em\lower.7ex\hbox{E}\kern-.125emX}}

\DeclareMathOperator{\sign}{sign}

\usepackage{fancyhdr}
\fancypagestyle{firstpage}{
  \fancyhf{}
  \fancyhead[C]{\small Extended version of the paper ``Latent Space Inpainting for Loss-Resilient Collaborative Object Detection,'' to be presented at the \textit{IEEE International Conference on Communications (ICC)}, Montreal, Canada, June 14-23, 2021.}
}

\begin{document}
\title{\vspace*{0.5cm} 
Latent-Space Inpainting for Packet Loss Concealment in Collaborative Object Detection
\thanks{This work was supported in part by the Natural Sciences and Engineering Research Council (NSERC) of Canada.}
}

\author{\vspace*{1cm}\IEEEauthorblockN{Ivan V. Baji\'{c}}\\
\IEEEauthorblockA{\textit{School of Engineering Science} \\
\textit{Simon Fraser University}\\
Burnaby, BC, Canada}
}

\maketitle

\doublespacing

\begin{abstract}
Edge devices, such as cameras and mobile units, are increasingly capable of performing sophisticated computation in addition to their traditional roles in sensing and communicating signals. The focus of this paper is on collaborative object detection, where deep features computed on the edge device from input images are transmitted to the cloud for further processing. We consider the impact of packet loss on the transmitted features and examine several ways for recovering the missing data. In particular, through theory and experiments, we show that methods for image inpainting based on partial differential equations work well for the recovery of missing features in the latent space. The obtained results represent the new state of the art for missing data recovery in collaborative object detection.   
\end{abstract}


\thispagestyle{firstpage}

\section{Introduction}
In video surveillance and monitoring systems, input video is usually sent to the cloud for temporary storage or further visual analysis. With the emergence of ``smart cameras,'' simpler forms of visual analysis can now be performed on-board, without the need to incur costs related to video transmission to the cloud or potential privacy breaches. Still, computational resources available in the cloud far outmatch those available at the edge, enabling much more sophisticated analysis in the cloud compared to what is possible in edge devices.

\begin{figure}
    \centering
    \includegraphics[width=0.5\columnwidth]{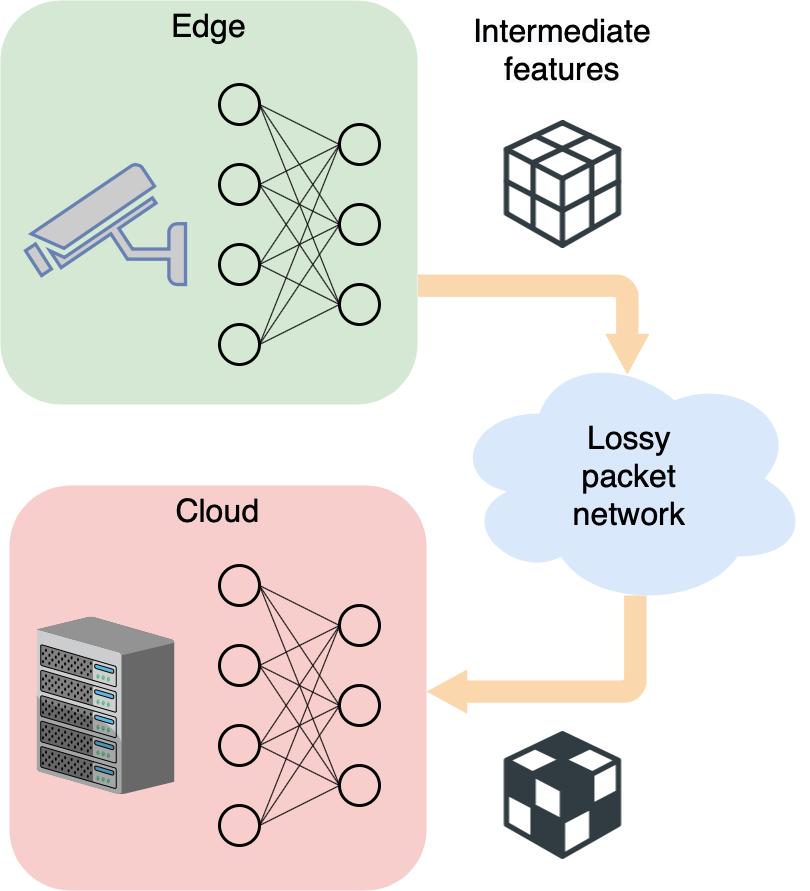}
    \caption{Collaborative intelligence over a lossy channel}
    \label{fig:CI_system}
\end{figure}

In between the two extremes mentioned above -- cloud-based and edge-based analytics -- stands \emph{collaborative intelligence} (CI)~\cite{kang2017neurosurgeon, eshratifar2019towards, Bajic_etal_ICASSP21}, a framework in which a machine learning model, usually a deep model, is split between the edge and the cloud. The front-end of the model is deployed at the edge and computes intermediate features, which are sent to the back-end in the cloud to complete the inference task, as shown in Fig.~\ref{fig:CI_system}. The CI approach has shown potential for better energy efficiency and lower latency compared to cloud-based or edge-based analytics~\cite{kang2017neurosurgeon}. Moreover, CI is better suited to privacy protection than the cloud-based approach, because the original input signal never leaves the edge device -- only the intermediate features are sent to the cloud.    

To make efficient use of the communication channel in CI, intermediate features need to be compressed. There has been increasing interest in feature compression  recently, both in the academic~\cite{dfc_for_collab_object_detection,eshratifar2019bottlenet,Choi2018NearLosslessDF,Chen19,Duan2020VideoCF,Saeed_ICIP19,Hyomin_ICASSP20,Saeed_ICASSP20,Bob_ICME20} and standardization community~\cite{MPEG_VCM_CFE}. Our focus here is on another aspect of the communication channel, namely its imperfections. At the network/transport layer, channel imperfections will manifest themselves as packet loss, leading to missing feature values. This is illustrated by dark regions in the feature tensor shown in Fig.~\ref{fig:CI_system}. For successful inference, this data loss in the feature tensors needs to be mitigated. Yet, there has been very limited amount of work on this topic. The authors in~\cite{choi_neural_2019, BottleNet++} studied joint source-channel coding of intermediate features to improve their robustness against bit errors. In~\cite{DFTS_2018}, simple interpolation methods (e.g., nearest neighbor and bi-linear) were explored for recovering missing features, while~\cite{Bragilevsky_Access_2020} proposed low-rank tensor completion for this purpose. All the aforementioned studies focused on image classification models.

In this paper we focus on the task of object detection, where the model needs to simultaneously localize multiple objects within an input image and classify each one of them. This is arguably more challenging than image classification, where one label per input image needs to be produced, and the corresponding detection models may be expected to be more sensitive to errors and feature loss than image classification models. Indeed, as will be seen in the results, object detection accuracy drops quickly as the packet loss increases, unless something is done to recover lost features: with only 5\% loss, detection accuracy drops by about 20\%, whereas the results in~\cite{DFTS_2018, Bragilevsky_Access_2020} show that for classification models, 5\% loss leads up to about 5\% loss in accuracy, depending on the model. Hence, missing feature recovery is crucial for collaborative object detection.

In order to recover lost features, we borrow an idea from image inpainting~\cite{bertalmio_image_2000}, specifically the Partial Differential Equation (PDE)-based inpainting~\cite{bertalmio_navier-stokes_2001,telea_image_2004,bertalmio_pde-based_2006}. Such methods operate by solving a PDE-based model of surface flow, as briefly described in Section~\ref{sec:preliminaries}. In order to understand what is the equivalent of image surface flow in the latent space of features, in Section~\ref{sec:ls_flow} we analyze the effects of typical operations found in deep convolutional models on the PDE describing the surface flow. The conclusion is that latent-space flow is described by the same PDE as the input flow, but with an appropriately scaled flow field. This analysis is put to the test in Section~\ref{sec:experiments}, where we compare the efficacy of latent-space inpainting against the current state-of-the-art for missing feature recovery. Finally, the paper is concluded in Section~\ref{sec:conclusions}.

\section{Preliminaries}
\label{sec:preliminaries}

\subsection{Surface flow}
Image inpainting is the problem of filling in the missing details of an image. PDE-based methods have been prominent in this area, initially developed as models that attempt to mimic the techniques used by professional painting restorators~\cite{bertalmio_image_2000}. Several PDE formulations have been used for this purpose, but our focus here is on a particular formulation from~\cite{bertalmio_navier-stokes_2001, bertalmio_pde-based_2006} that we will refer to as ``surface flow.''  This formulation allows image surface to ``flow'' into the missing area as the inpainting progresses. Let $I(x,y,t)$ denote pixel intensity at time $t$, at spatial position $(x,y)$, then the surface flow can be expressed as~\cite{bertalmio_navier-stokes_2001, bertalmio_pde-based_2006}: 
\begin{equation}
    \frac{\partial I}{\partial x}v_x + \frac{\partial I}{\partial y}v_y + \frac{\partial I}{\partial t} = 0,
    \label{eq:2D+t_surface_flow}
\end{equation}
where $(v_x,v_y)$ represents the flow vector. 
Here, spatial coordinates $(x,y)$ and iterations $t$ are expressed as continuous quantities to allow PDE-based formulation, but in practice they are discrete. We also note that~(\ref{eq:2D+t_surface_flow}) is the same equation as optical flow~\cite{Horn_Schunk_1981}, but there, $(v_x,v_y)$ represents motion between video frames. This analogy is not surprising, because both surface flow and optical flow represent conservation laws for image intensity. In fact, latent space motion analysis based on optical flow has recently been performed in~\cite{Mateen_ICASSP21}. In Section~\ref{sec:ls_flow} we will explore what happens to~(\ref{eq:2D+t_surface_flow}) as the image $I$ passes through common operations found in deep convolutional models.

\subsection{Feature packetization}
In order to transmit the feature tensor produced by the edge sub-model over a packet network, tensor elements need to be packetized. There is currently no ``standard'' way of doing this, and there are many possible ways of forming packets. For the purposes of this study, we adopt the following approach. First, tensor channels are tiled into an image, as in~\cite{dfc_for_collab_object_detection}. Then packets are formed by taking eight rows at a time from such a tiled image. This is similar to how packets are usually formed in video streaming~\cite{Wang_etal_2002}, and also resembles the way packets are formed from feature tensors in~\cite{DFTS_2018,Bragilevsky_Access_2020}. With such a packetization scheme, each lost packet creates an 8-row gap in the channels of the feature tensor, which needs to be filled in. An example will be seen in Fig.~\ref{fig:visual_examples} in Section~\ref{sec:experiments}. Note that the analysis in Section~\ref{sec:ls_flow} is independent of the packetization scheme and none of the feature recovery methods examined here crucially depend on it. However, numerical results may change if a different packetization scheme is used. 

\section{Latent-space surface flow}
\label{sec:ls_flow}
\begin{figure}
    \centering
    \includegraphics[width=0.5\columnwidth]{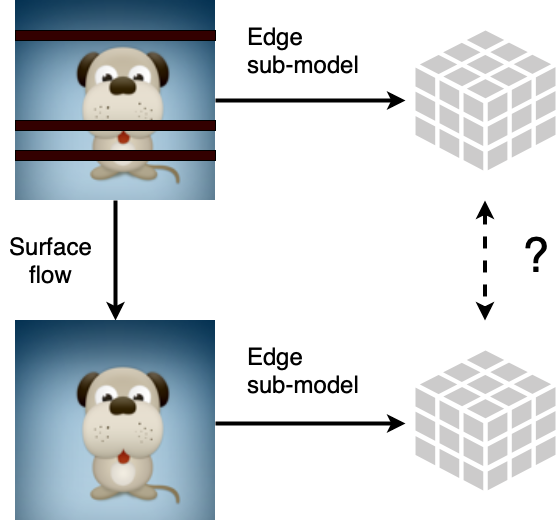}
    \caption{What is the latent-space equivalent of input-space surface flow?}
    \label{fig:overview}
\end{figure}

The success of PDE-based image inpainting has demonstrated that~(\ref{eq:2D+t_surface_flow}) represents a good model for natural images, so it can be used to fill in the gaps in such images. If that is the case, what would be a good analogous model for latent-space feature tensors, whose gaps we need to fill in? This is not immediately clear, since latent-space feature tensors may look quite different from natural images, as will be seen in Fig.~\ref{fig:visual_examples}. In this section, our goal is to find the latent-space equivalent of input-space surface flow, as illustrated in Fig.~\ref{fig:overview}. 

To do this, we look at the processing  pipeline between the input image and a given channel of a feature tensor in an edge sub-model. In most deep convolutional networks, this processing pipeline consists of a sequence of basic operations: convolutions, nonlinear pointwise activations, and pooling. We will analyze the effect of each of these operations on~(\ref{eq:2D+t_surface_flow}).

\subsection{Convolution}
\label{sec:convolution}
Let $f$ be a (spatial) filter kernel, then the surface flow after convolution can be described by
\begin{equation}
    \frac{\partial}{\partial x}(f*I)v'_x  + \frac{\partial}{\partial y}(f*I)v'_y + \frac{\partial}{\partial t}(f*I) = 0,
    \label{eq:2D+t_conv}
\end{equation}
where $*$ represents convolution and $(v'_x,v'_y)$ is the flow after the convolution. Since the convolution and differentiation are linear operations, we have
\begin{equation}
    f*\left(\frac{\partial I}{\partial x}v'_x  +
    \frac{\partial I}{\partial y}v'_y + 
    \frac{\partial I}{\partial t} \right) = 0.
    \label{eq:2D+t_conv_2}
\end{equation}
Hence, a solution to the surface flow after the convolution satisfies the same type of equation as~(\ref{eq:2D+t_surface_flow}). This means that if we had a method for solving~(\ref{eq:2D+t_surface_flow}), the same method would be able to find a solution to the post-convolution flow~(\ref{eq:2D+t_conv_2}). 

\subsection{Nonlinear activation} 
Nonlinear activations such as \texttt{sigmoid}, \texttt{tanh}, \texttt{ReLU}, etc., are usually applied on the output of convolutions in deep models~\cite{Goodfellow-et-al-2016}. These are point-wise operations, applied to each sample separately. Let $\sigma(\cdot)$ denote such a nonlinear activation, then the surface flow after this operation is described by the following PDE:
\begin{equation}
    \frac{\partial}{\partial x}\left[\sigma(I)\right]v'_x  + \frac{\partial}{\partial y}\left[\sigma(I)\right]v'_y  + \frac{\partial}{\partial t}\left[\sigma(I)\right] = 0,
    \label{eq:2D+t_nonlin}
\end{equation}
where $(v'_x,v'_y)$ is the flow after the nonlinear activation. By using the chain rule of differentiation, the above equation can be rewritten as
\begin{equation}
    \sigma'(I)\cdot \left(\frac{\partial I}{\partial x}v'_x  +
    \frac{\partial I}{\partial y}v'_y  +
    \frac{\partial I}{\partial t} \right) = 0.
    \label{eq:2D+t_nonlin_2}
\end{equation}
Hence, again, $(v'_x,v'_y)$ satisfies the same type of equation as~(\ref{eq:2D+t_surface_flow}), and a method that solves pre-activation flow~(\ref{eq:2D+t_surface_flow}) should be able to find a solution to post-activation flow~(\ref{eq:2D+t_nonlin_2}). 

Note that~(\ref{eq:2D+t_nonlin_2}) may also have other solutions, not just those that satisfy the surface flow equation~(\ref{eq:2D+t_surface_flow}). For example, consider the \texttt{ReLU} activation defined by $\sigma(x)=\max(0,x)$. In the regions of $I$ where the values are are negative, the corresponding outputs of \texttt{ReLU} will be zero, so $\sigma(I)=0$, and its derivative is also zero, $\sigma'(I)=0$. Hence, in those regions,~(\ref{eq:2D+t_nonlin_2}) can be satisfied for arbitrary flow $(v'_x,v'_y)$. However, in practice this does not matter, because in regions where the signal is constant, propagating signal values in any direction will produce the same result.  

\subsection{Pooling} 
Various forms of pooling are used in deep models~\cite{Goodfellow-et-al-2016}, such as max-pooling, mean-pooling, learnt pooling via strided convolutions, etc. All these forms of pooling can be decomposed into a sequence of two operations as follows
$$\underbrace{\text{spatial operation} \to \text{scale change}}_{\text{pooling}}$$
In the case of mean pooling or learnt pooling, spatial operation is a convolution. In the case of max-pooling, spatial operation is a local maximum operation. Scale change is simply implemented using regular downsampling. Since the effect of convolution on surface flow was discussed in Section~\ref{sec:convolution}, here we further discuss the effect of local maximum and scale change on~(\ref{eq:2D+t_surface_flow}). 

\subsubsection{Local maximum} 
Consider the maximum of function $I(x,y,t)$ over a local spatial region $$R_h=[x_0-h, x_0+h]\times[y_0-h, y_0+h],$$ at a given time $t$. We can approximate $I(x,y,t)$ over this region as a locally-linear function, whose slope is determined by the spatial derivatives of $I$ at $(x_0,y_0)$, namely $\frac{\partial}{\partial x}I$ and $\frac{\partial}{\partial y}I$. Depending on the sign of these derivatives, the local maximum of $I$ over $R_h$ will be somewhere on the boundary of $R_h$. In the special case where both derivatives are zero, $I$ is constant over $R_h$ and any point in $R_h$, including boundary points, is a local maximum. 

From the first-order Taylor series expansion of $I(x,y,t)$ around $(x_0,y_0,t)$ we have
\begin{equation}
\begin{split}
    I(x_0 + \epsilon_x,y_0 + \epsilon_y,t) \approx \: & I(x_0,y_0,t) \\ & +  \frac{\partial}{\partial x}I(x_0,y_0,t)\cdot \epsilon_x \\ &+ \frac{\partial}{\partial y}I(x_0,y_0,t)\cdot \epsilon_y,
\end{split}
\end{equation}
for $|\epsilon_x|,|\epsilon_y| \leq h$. With such linear approximation, the local maximum of $I(x,y,t)$ over $R_h$ (which, as we saw above, is somewhere on the boundary of $R_h$) can be approximated as
\begin{equation}
\begin{split}
    \max_{(x,y)\in R_h} &  I(x,y,t)  \approx  I(x_0,y_0,t) \\
    & + \sign\left(\frac{\partial}{\partial x}I(x_0,y_0,t)\right) \cdot \frac{\partial}{\partial x}I(x_0,y_0,t) \cdot h \\
    & + \sign\left(\frac{\partial}{\partial y}I(x_0,y_0,t)\right) \cdot \frac{\partial}{\partial y}I(x_0,y_0,t) \cdot h.
\end{split}
\label{eq:local_max}
\end{equation}
Let~(\ref{eq:local_max}) be the definition of $M(x_0,y_0,t)$, the function that takes on local maximum values of $I(x,y,t)$ over spatial windows of size $2h\times2h$. The surface flow after such a local maximum operation is described by
\begin{equation}
    \frac{\partial M}{\partial x}v'_x + \frac{\partial M}{\partial y}v'_y  + \frac{\partial M}{\partial t} = 0,
    \label{eq:2D+t_local_max}
\end{equation}
where $(v'_x,v'_y)$ represents the flow after local spatial maximum operation. From~(\ref{eq:local_max}) we have 
\begin{equation}
\begin{aligned}
    \frac{\partial M}{\partial x} &= \frac{\partial I}{\partial x} + \sign\left(\frac{\partial I}{\partial x}\right)\frac{\partial^2 I}{\partial x^2}h + \sign\left(\frac{\partial I}{\partial y}\right)\frac{\partial^2 I}{\partial x \partial y}h,\\
    \frac{\partial M}{\partial y} &= \frac{\partial I}{\partial y} + \sign\left(\frac{\partial I}{\partial x}\right)\frac{\partial^2 I}{\partial x \partial y}h + \sign\left(\frac{\partial I}{\partial y}\right)\frac{\partial^2 I}{\partial y^2}h, \\
    \frac{\partial M}{\partial t} &= \frac{\partial I}{\partial t} + \sign\left(\frac{\partial I}{\partial x}\right)\frac{\partial^2 I}{\partial x \partial t}h + \sign\left(\frac{\partial I}{\partial y}\right)\frac{\partial^2 I}{\partial y \partial t}h,
\end{aligned}
\label{eq:partials_max}
\end{equation}
and substituting~(\ref{eq:partials_max}) in~(\ref{eq:2D+t_local_max}) gives the following PDE
\begin{equation}
\begin{split}
    \frac{\partial I}{\partial x}v'_x  &+ \frac{\partial I}{\partial y}v'_y + \frac{\partial I}{\partial t} \\ &+  \sign\left(\frac{\partial I}{\partial x}\right) \cdot \frac{\partial}{\partial x} \left( \frac{\partial I}{\partial x}v'_x + \frac{\partial I}{\partial y}v'_y  + \frac{\partial I}{\partial t}\right)\cdot h \\ &+
    \sign\left(\frac{\partial I}{\partial y}\right) \cdot \frac{\partial}{\partial y} \left( \frac{\partial I}{\partial x}v'_x + \frac{\partial I}{\partial y}v'_y  + \frac{\partial I}{\partial t}\right)\cdot h = 0.
\end{split}
\label{eq:2D+t_post_max}
\end{equation}
Note that if 
\begin{equation}
    \frac{\partial I}{\partial x}v'_x + \frac{\partial I}{\partial y}v'_y  + \frac{\partial I}{\partial t}=0,
    \label{eq:new_flow}
\end{equation}
then~(\ref{eq:2D+t_post_max}) will automatically be satisfied. But~(\ref{eq:2D+t_post_max}) is the same PDE as~(\ref{eq:2D+t_surface_flow}), with $(v'_x,v'_y)$ now playing the role of $(v_x,v_y)$.  
Hence, if $(v'_x,v'_y)$ satisfies the same type of surface flow equation as~(\ref{eq:2D+t_surface_flow}), it will also satisfy~(\ref{eq:2D+t_post_max}). So a method that solves~(\ref{eq:2D+t_surface_flow}) should be able to find, at least approximately, the surface flow after the local maximum operation.

\subsubsection{Scale change} 
Finally, consider the change of spatial scale by factors $s_x$ and $s_y$ in x- and y-directions, such that the new signal is $I'(x,y,t) =I(s_x\cdot x,s_y\cdot y,t)$. The surface flow equation after spatial scaling is
\begin{equation}
    \frac{\partial I'}{\partial x}v'_x  + \frac{\partial I'}{\partial y}v'_y + \frac{\partial I'}{\partial t} = 0.
    \label{eq:2D+t_scale_change}
\end{equation}
Since $\frac{\partial I'}{\partial x}=s_x\cdot \frac{\partial I}{\partial x}$, $\frac{\partial I'}{\partial y}=s_y\cdot \frac{\partial I}{\partial y}$, and $\frac{\partial I'}{\partial t}=\frac{\partial I}{\partial t}$, it is easy to see that post-scaling flow satisfies the same equation as~(\ref{eq:2D+t_surface_flow}), but with the correspondingly scaled flow field: $(v'_x,v'_y)=(v_x/s_x,v_y/s_y)$, where $(v_x,v_y)$ is the pre-scaling flow. In deep convolutional models, scaling factors of $s_x=s_y=2$ are commonly found, so the surface flow in the downscaled signal is correspondingly scaled by a factor of $2$ as well. 

The analysis presented above suggests that the surface flow equation is largely left intact by the common operations found in deep convolutional models, such as convolutions, nonlinear activations, and pooling. In the case of max-pooling, the conclusion is only approximate, but over small windows such as $2\times2$, which are common, it is expected to be a good approximation. Hence, a method that solves the surface flow PDE~(\ref{eq:2D+t_surface_flow}) should be a good solution to the latent-space surface flow as well. We put this conclusion to the test in the next section, where we deploy two algorithms for finding surface flow, from~\cite{bertalmio_navier-stokes_2001} and~\cite{telea_image_2004}, to recover missing features in the latent space.

\section{Experiments}
\label{sec:experiments}

\subsection{Setup}
The experiments are carried out on the YOLOv3 object detector~\cite{redmon2018yolov3}. Specifically, a Python implementation of this model\footnote{\texttt{https://github.com/experiencor/keras-yolo3}} based on Keras and Tensorflow was used. Details of the experimental testbed are shown in Table~\ref{tab:exp_testbed}. YOLOv3 has a complex architecture\footnote{\texttt{https://towardsdatascience.com/yolo-v3-object-detection-53fb7d3bfe6b}} with a number of skip connections. Depending on where the model is split, one or more tensors needs to be transmitted. If one wishes to transmit only a single tensor per input image, there  are two points where a sensible split can be made - layer 12 and layer 36. In this paper we picked the deeper of these two split points, namely layer 36, so that layers 1-36 form the edge sub-model in Fig.~\ref{fig:CI_system} and the remaining layers 37-105 constitute the cloud sub-model.  

Input images were resized to $512 \times 512$ before feeding them to the edge sub-model. With this input size, the feature tensor produced by the edge sub-model is $64 \times 64 \times 256$, i.e., $256$ channels of size $64 \times 64$. These were tiled into a square image of size $1024 \times 1024$, and quantized to 8-bit precision, similarly to~\cite{dfc_for_collab_object_detection,Choi2018NearLosslessDF}. No further compression of feature values was employed. 

\begin{table}[tbp]
\caption{Experimental testbed}
\begin{center}
\begin{tabular}{|c|l|}
\hline
CPU & Intel(R) Core(TM) i7-6700K @ 4.00GHz\\
\hline
GPU &  NVIDIA GeForce GTX Titan X \\
\hline
CUDA &  10.2 \\
\hline
RAM & 64 GB\\
\hline
Oprating system & Ubuntu 18.04 LTS\\
\hline
Language &  Python 3.5.6\\
\hline
Tensorflow &  2.2.0 \\
\hline
Keras &  2.3.1 \\
\hline
OpenCV &  3.1.0 \\
\hline
\end{tabular}
\label{tab:exp_testbed}
\end{center}
\end{table}

We used two methods for solving surface flow~(\ref{eq:2D+t_surface_flow}): one, from~\cite{bertalmio_navier-stokes_2001}, will be referred to as ``Navier-Stokes,'' and the other, from~\cite{telea_image_2004}, will be referred to as ``Telea.'' These methods were compared against the current state of the art in missing feature recovery~\cite{Bragilevsky_Access_2020}. In~\cite{Bragilevsky_Access_2020}, several tensor completion methods were used to recover tensor features missing due to packet loss. These tensor completion methods make no assumptions on how the feature tensor is produced. The only underlying assumption is that the tensor lies in a low-rank manifold which the completion method tries to reach. In~\cite{Bragilevsky_Access_2020}, tensor completion methods were tested on recovering features from image classification models VGG-16~\cite{VGG} and ResNet-34~\cite{ResNet}, and found to offer relatively similar performance. As a representative of these methods, we use Simple Low-Rank Tensor Completion (SiLRTC)~\cite{LRTC}. It is an iterative method that refines its estimates of missing values at each iteration. 

\subsection{Quantitative results}
\begin{figure}
    \centering
    \includegraphics[trim=10 0 30 0,clip,width=0.6\columnwidth]{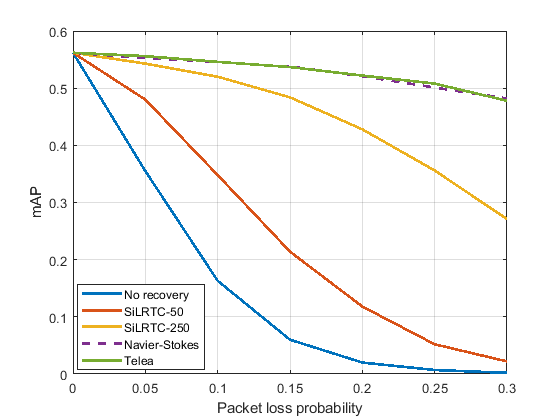}
    \caption{mAP vs. loss for various feature recovery methods}
    \label{fig:mAP_vs_loss}
\end{figure}

Recovery of missing features was tested on the 2017 version of the validation set from the COCO dataset~\cite{COCO}. This set contains 5,000 images with 80 object classes represented in the set. Independent packet loss was simulated with loss probabilities $p \in \{0.05, 0.10, 0.15, 0.20, 0.25, 0.30\}$. For each input image, the loss was applied to the tiled feature tensor, and the missing values were recovered by various methods. Model accuracy was measured in terms of the mean Average Precision (mAP)~\cite{redmon2018yolov3} at the Intersection over Union (IoU) threshold of 0.5. These are common settings in the literature for quantifying object detection accuracy. The results are shown in Fig.~\ref{fig:mAP_vs_loss}.   

In the figure, the blue line represents the case where no feature recovery is performed, and the missing values are simply set to zero. As shown by this curve, the accuracy of object detection quickly drops from its loss-free value of 0.56 down to about 0.36 with just 5\% loss. The curve labeled SiLRTC-50 is the performance obtained when missing features are recovered using SiLRTC with 50 iterations. This provides some improvement. Much better accuracy is obtained when running SiLRTC with 250 iterations, as shown by the curve labeled SiLRTC-250. It should be noted that in~\cite{Bragilevsky_Access_2020}, 50 iterations of SiLRTC were found to be sufficient to provide solid accuracy for VGG-16 and ResNet-34, whereas here, with YOLOv3, a larger number of iterations is needed. One factor that may play a role in this is tensor size; here we are dealing with tensors of size $64 \times 64 \times 256$, whereas in~\cite{Bragilevsky_Access_2020}, tensors were much smaller: $14 \times 14 \times 512$ for VGG-16 and $28 \times 28 \times 128$ for ResNet-34.    

The best accuracy in Fig.~\ref{fig:mAP_vs_loss} is obtained by the two PDE-based inpainting methods, Navier-Stokes and Telea. Their performance in the figure is virtually indistinguishable, which is not surprising considering that they are based on the same principles. As seen in the figure, PDE-based inpainting methods are able to achieve significant improvement in object detection accuracy over tensor completion methods. Part of the reason for this is that they reasonably manage to capture the structure of the tensor via surface flow~(\ref{eq:2D+t_surface_flow}), whereas tensor completion methods do not use any such insight and try to discover this structure iteratively. 

In the second column of Table~\ref{tab:results} we summarize the average mAP gain from Fig.~\ref{fig:mAP_vs_loss} of various methods over the case of no recovery. These values are obtained by numerically integrating the area between the ``No recovery'' curve and the curve corresponding to a particular method using the trapezoidal rule, and then dividing this value by the range of packet loss probabilities, which is 0.3. Hence, these values represents the average mAP gain over this range of packet loss probabilities. As seen in the table, the two PDE-based inpainting methods provide significant average mAP improvement of over 0.38. SiLRTC is also able to provide solid gains of about 0.31, if executed for 250 iterations.

\begin{figure*}[t]
    \centering
    \includegraphics[width=\textwidth]{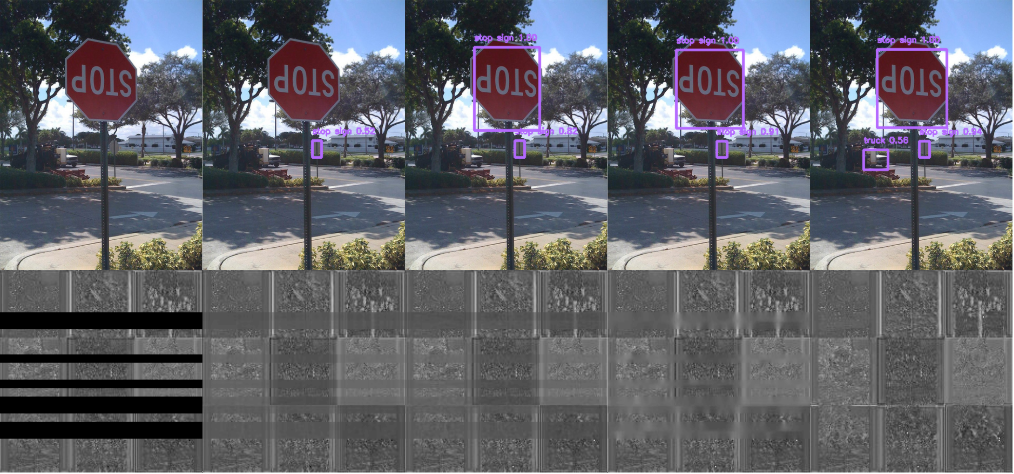}
    \caption{\textbf{Top row:} object detections overlaid on the original image. \textbf{Bottom row:} part of the latent space (9 tensor channels) corresponding to the object detections in the top row. Tensor channels have been mapped to grayscale and enhanced for visualization purposes. From left to right: (1) Without any data recovery (locations of lost data indicated as black), no detections are made; SiLRTC-50 (not shown) also does not provide sufficient recovery for any detections to be made. (2) With SiLRTC-150, one STOP sign is detected with confidence 0.52. (3) With SiLRTC-250, two STOP signs are detected, and the confidence for the smaller sign is improved to 0.82. (4) With Navier-Stokes~\cite{bertalmio_navier-stokes_2001} and Telea~\cite{telea_image_2004} inpainting, two STOP signs are detected, and the confidence for the smaller sign is improved to 0.91. (5) Without any loss, a truck is also detected, and the confidence for the smaller STOP sign is 0.94. }
    \label{fig:visual_examples}
    \vspace{-3pt}
\end{figure*}

The last column in Table~\ref{tab:results} shows the average execution time per tensor of various methods. As seen here, the PDE-based methods are not only more accurate, but much faster than SiLRTC - over 100$\times$ faster than SiLRTC-50 and almost 600$\times$ faster than SiLRTC-250. The reason for this is that SiLRTC performs a global decomposition of the tensor using Singular Value Decomposition (SVD), which is quite expensive computationally. Meanwhile, the PDE-based inpainting methods operate locally. While it might be possible for SiLRTC to reach the accuracy provided by the PDE-based methods using more iterations, this would run counter to the goals of collaborative intelligence, where latency is an important aspect; with 250 iterations, it is already much slower than the PDE-based methods.

\begin{table}[tbp]
\caption{Average mAP gain and execution time of various feature recovery methods}
\begin{center}
\begin{tabular}{|c|c|r|}
\hline
\textbf{Method} & \textbf{Avg. mAP gain} & \textbf{Time per tensor (sec.)} \\
\hline
\hline
SiLRTC-50 & 0.1028 &  17.0793\hspace{0.9cm} \\
\hline
SiLRTC-250 & 0.3101 &  83.2044\hspace{0.9cm} \\
\hline
Navier-Stokes & 0.3823 & 0.1408\hspace{0.9cm} \\
\hline
Telea & 0.3837 &  0.1356\hspace{0.9cm} \\
\hline
\end{tabular}
\label{tab:results}
\end{center}
\end{table}

\subsection{Visual examples}
Finally, we show several visual examples in Fig.~\ref{fig:visual_examples}. The top row in the figure shows object detections overlaid on the original image, and the bottom row shows nine channels from the feature tensor produced by the edge sub-model. Note that the entire tensor contains 256 channels, so the nine channels in the figure are only a small part of the latent space. 

The leftmost images in the figure correspond to the case where no feature recovery is performed. One can see the black lines in the tensor channels, indicating the locations of the missing packets. No objects are detected when such a tensor is fed to the cloud sub-model. The next case to the right is SiLRTC-50, which manages to provide some recovery to the missing features, resulting in the detection of a small STOP sign in the background, which is barely detected with confidence of 0.52. In the next case to the right, SiLRTC-250 provides somewhat better recovery and the resulting tensor allows detection of the large STOP sign in the foreground, as well as the small sign in the background, now with increased confidence of 0.82. Finally, Telea inpainting provides even better feature recovery, with increased confidence of 0.91 for the small STOP sign. the Navier-Stokes inpainting gave almost the same results. The rightmost image corresponds to  the case when there are no lost features; in this case, the confidence about the small STOP sign is 0.94, and a small truck is also detected in the background. 

\section{Conclusions}
\label{sec:conclusions}
In this paper, the problem of missing feature recovery in collaborative object detection was studied. Starting with the surface flow model of natural images, which is known to work well for image inpainting, we analyzed the effect of various processing steps used in deep convolutional models on such flow, and concluded that the flow equation remains approximately intact under such operations. Hence, methods that work well for image inpainting should work well for latent-space inpainting as well. This conclusion was tested on the YOLOv3 object detector. The results showed that the resulting latent-space inpainting methods provide significant improvement over existing tensor completion-based methods for missing feature recovery, in both accuracy and speed. 



\bibliographystyle{IEEEtran}
\bibliography{ref}

\begin{thebibliography}{10}
\providecommand{\url}[1]{#1}
\csname url@samestyle\endcsname
\providecommand{\newblock}{\relax}
\providecommand{\bibinfo}[2]{#2}
\providecommand{\BIBentrySTDinterwordspacing}{\spaceskip=0pt\relax}
\providecommand{\BIBentryALTinterwordstretchfactor}{4}
\providecommand{\BIBentryALTinterwordspacing}{\spaceskip=\fontdimen2\font plus
\BIBentryALTinterwordstretchfactor\fontdimen3\font minus
  \fontdimen4\font\relax}
\providecommand{\BIBforeignlanguage}[2]{{%
\expandafter\ifx\csname l@#1\endcsname\relax
\typeout{** WARNING: IEEEtran.bst: No hyphenation pattern has been}%
\typeout{** loaded for the language `#1'. Using the pattern for}%
\typeout{** the default language instead.}%
\else
\language=\csname l@#1\endcsname
\fi
#2}}
\providecommand{\BIBdecl}{\relax}
\BIBdecl

\bibitem{kang2017neurosurgeon}
Y.~Kang, J.~Hauswald, C.~Gao, A.~Rovinski, T.~Mudge, J.~Mars, and L.~Tang,
  ``Neurosurgeon: Collaborative intelligence between the cloud and mobile
  edge,'' in \emph{Proc. 22nd ACM Int. Conf. Arch. Support Programming
  Languages and Operating Syst.}, 2017, pp. 615--629.

\bibitem{eshratifar2019towards}
A.~E. Eshratifar, A.~Esmaili, and M.~Pedram, ``Towards collaborative
  intelligence friendly architectures for deep learning,'' in \emph{20th Int.
  Symp. Quality Electronic Design (ISQED)}, 2019, pp. 14--19.

\bibitem{Bajic_etal_ICASSP21}
I.~V. Baji\'{c}, W.~Lin, and Y.~Tian, ``Collaborative intelligence: Challenges
  and opportunities,'' in \emph{Proc. IEEE ICASSP}, 2021, to appear.

\bibitem{dfc_for_collab_object_detection}
H.~Choi and I.~V. Baji\'{c}, ``Deep feature compression for collaborative
  object detection,'' in \emph{Proc. IEEE ICIP}, 2018, pp. 3743--3747.

\bibitem{eshratifar2019bottlenet}
A.~E. Eshratifar, A.~Esmaili, and M.~Pedram, ``{BottleNet:} a deep learning
  architecture for intelligent mobile cloud computing services,'' in \emph{2019
  IEEE/ACM Int. Symp. Low Power Electr. Design (ISLPED)}, 2019.

\bibitem{Choi2018NearLosslessDF}
H.~Choi and I.~V. Baji\'{c}, ``Near-lossless deep feature compression for
  collaborative intelligence,'' in \emph{Proc. IEEE MMSP}, 2018, pp. 1--6.

\bibitem{Chen19}
Z.~Chen, K.~Fan, S.~Wang, L.~Duan, W.~Lin, and A.~C. Kot, ``Toward intelligent
  sensing: Intermediate deep feature compression,'' \emph{IEEE Trans. Image
  Processing}, vol.~29, pp. 2230--2243, 2019.

\bibitem{Duan2020VideoCF}
L.~Duan, J.~Liu, W.~Yang, T.~Huang, and W.~Gao, ``Video coding for machines: A
  paradigm of collaborative compression and intelligent analytics,'' \emph{IEEE
  Trans. Image Processing}, vol.~29, pp. 8680--8695, 2020.

\bibitem{Saeed_ICIP19}
S.~R. Alvar and I.~V. Baji\'{c}, ``Multi-task learning with compressible
  features for collaborative intelligence,'' in \emph{Proc. IEEE ICIP}, Sep.
  2019, pp. 1705--1709.

\bibitem{Hyomin_ICASSP20}
H.~Choi, R.~A. Cohen, and I.~V. Baji\'{c}, ``Back-and-forth prediction for deep
  tensor compression,'' in \emph{Proc. IEEE ICASSP}, 2020, pp. 4467--4471.

\bibitem{Saeed_ICASSP20}
S.~R. Alvar and I.~V. Baji\'{c}, ``Bit allocation for multi-task collaborative
  intelligence,'' in \emph{Proc. IEEE ICASSP}, May 2020, pp. 4342--4346.

\bibitem{Bob_ICME20}
R.~A. Cohen, H.~Choi, and I.~V. Baji\'{c}, ``Lightweight compression of neural
  network feature tensors for collaborative intelligence,'' in \emph{Proc. IEEE
  ICME}, Jul. 2020, pp. 1--6.

\bibitem{MPEG_VCM_CFE}
ISO/IEC, ``Draft call for evidence for video coding for machines,'' Jul. 2020,
  {ISO/IEC JTC 1/SC 29/WG 11 W19508}.

\bibitem{choi_neural_2019}
K.~Choi, K.~Tatwawadi, A.~Grover, T.~Weissman, and S.~Ermon, ``Neural joint
  source-channel coding,'' in \emph{Proc. {ICML}}, Jun. 2019, pp. 1182--1192.

\bibitem{BottleNet++}
J.~{Shao} and J.~{Zhang}, ``Bottlenet++: An end-to-end approach for feature
  compression in device-edge co-inference systems,'' in \emph{Proc. IEEE ICC
  Workshops}, 2020, pp. 1--6.

\bibitem{DFTS_2018}
H.~Unnibhavi, H.~Choi, S.~R. Alvar, and I.~V. Baji\'{c}, ``{DFTS}: Deep feature
  transmission simulator,'' in \emph{IEEE MMSP}, 2018, demo.

\bibitem{Bragilevsky_Access_2020}
L.~{Bragilevsky} and I.~V. Baji\'{c}, ``Tensor completion methods for
  collaborative intelligence,'' \emph{IEEE Access}, vol.~8, pp.
  41\,162--41\,174, 2020.

\bibitem{bertalmio_image_2000}
M.~Bertalmio, G.~Sapiro, V.~Caselles, and C.~Ballester, ``Image inpainting,''
  in \emph{Proc. {SIGGRAPH}}, Jul. 2000, pp. 417--424.

\bibitem{bertalmio_navier-stokes_2001}
M.~Bertalmio, A.~L. Bertozzi, and G.~Sapiro, ``{Navier-Stokes}, fluid dynamics,
  and image and video inpainting,'' in \emph{Proc. IEEE CVPR}, vol.~1, Dec.
  2001, pp. I--355--I--362.

\bibitem{telea_image_2004}
A.~Telea, ``An image inpainting technique based on the fast marching method,''
  \emph{J. Graphics Tools}, vol.~9, no.~1, pp. 23--34, Jan. 2004.

\bibitem{bertalmio_pde-based_2006}
M.~Bertalmío, V.~Caselles, G.~Haro, and G.~Sapiro, ``{PDE}-based image and
  surface inpainting,'' in \emph{Handbook of {Mathematical} {Models} in
  {Computer} {Vision}}, N.~Paragios, Y.~Chen, and O.~Faugeras, Eds.\hskip 1em
  plus 0.5em minus 0.4em\relax Boston, MA: Springer US, 2006, pp. 33--61.

\bibitem{Horn_Schunk_1981}
B.~K.~P. Horn and B.~G. Schunck, ``Determining optical flow,'' \emph{Artificial
  Intelligence}, vol.~17, no.~1, pp. 185 -- 203, 1981.

\bibitem{Mateen_ICASSP21}
M.~Ulhaq and I.~V. Baji\'{c}, ``Latent space motion analysis for collaborative
  intelligence,'' in \emph{Proc. IEEE ICASSP}, 2021, to appear.

\bibitem{Wang_etal_2002}
Y.~Wang, J.~Ostermann, and Y.-Q. Zhang, \emph{Video Processing and
  Communications}.\hskip 1em plus 0.5em minus 0.4em\relax Upper Saddle River,
  NJ, USA: Prentice-Hall, 2002.

\bibitem{Goodfellow-et-al-2016}
I.~Goodfellow, Y.~Bengio, and A.~Courville, \emph{Deep Learning}.\hskip 1em
  plus 0.5em minus 0.4em\relax MIT Press, 2016.

\bibitem{redmon2018yolov3}
J.~Redmon and A.~Farhadi, ``{YOLOv3:} an incremental improvement,'' \emph{arXiv
  preprint arXiv:1804.02767}, 2018.

\bibitem{VGG}
K.~Simonyan and A.~Zisserman, ``Very deep convolutional networks for
  large-scale image recognition,'' in \emph{Proc. ICLR}, May 2015.

\bibitem{ResNet}
K.~{He}, X.~{Zhang}, S.~{Ren}, and J.~{Sun}, ``Deep residual learning for image
  recognition,'' in \emph{Proc. IEEE CVPR}, Jun. 2016, pp. 770--778.

\bibitem{LRTC}
J.~{Liu}, P.~{Musialski}, P.~{Wonka}, and J.~{Ye}, ``Tensor completion for
  estimating missing values in visual data,'' \emph{IEEE Trans. Pattern
  Analysis and Machine Intelligence}, vol.~35, no.~1, pp. 208--220, Jan. 2013.

\bibitem{COCO}
T.-Y. Lin, M.~Maire, S.~Belongie, J.~Hays, P.~Perona, D.~Ramanan,
  P.~Doll{\'a}r, and C.~L. Zitnick, ``Microsoft {COCO:} common objects in
  context,'' in \emph{Proc. ECCV}, 2014, pp. 740--755.

\end{thebibliography}

\end{document}